\documentclass[10pt,twocolumn,letterpaper]{article}
\usepackage{cvpr}
\usepackage{times}
\usepackage{epsfig}
\usepackage{graphicx}
\usepackage{amsmath}
\usepackage{amssymb}

\usepackage{acronym}
\usepackage{subfig}
\usepackage{xfrac}
\usepackage{multirow}
\usepackage{dblfloatfix}
\usepackage{authblk}

\graphicspath{{Images/}}

\acrodef{SLAM}{Simultaneous Localisation and Mapping}
\acrodef{VO}{Visual Odometry}
\acrodef{SPP}{Spatial Pooling Pyramid}
\acrodef{SAND}{Scale-Adaptive Neural Dense}
\acrodef{AUC}{Area Under the Curve}

\def\p#1{\boldsymbol{p}_#1} 
\def\F#1{\boldsymbol{F}_#1} 
\def\C#1{\boldsymbol{c}_#1} 

\usepackage[pagebackref=true,breaklinks=true,letterpaper=true,colorlinks,bookmarks=false]{hyperref}

\cvprfinalcopy 

\def\cvprPaperID{3177} 

\ifcvprfinal\fi

\begin{document}
\title{\vspace{-0.5cm}Scale-Adaptive Neural Dense Features: Learning via Hierarchical Context Aggregation}

\author{\vspace{-0.5cm}Jaime Spencer,\quad Richard Bowden,\quad Simon Hadfield\\  \vspace{-0.2cm}
Centre for Vision, Speech and Signal Processing (CVSSP)\\
University of Surrey\\
{\tt\small \{jaime.spencer, r.bowden, s.hadfield\}@surrey.ac.uk}
}

\maketitle
\thispagestyle{empty}

\begin{abstract}
\vspace{-0.4cm}
How do computers and intelligent agents view the world around them?
Feature extraction and representation constitutes one the basic building blocks towards answering this question. 
Traditionally, this has been done with carefully engineered hand-crafted techniques such as HOG, SIFT or ORB.
However, there is no ``one size fits all'' approach that satisfies all requirements.

In recent years, the rising popularity of deep learning has resulted in a myriad of end-to-end solutions to many computer vision problems. 
These approaches, while successful, tend to lack scalability and can't easily exploit information learned by other systems. 

Instead, we propose SAND features, a dedicated deep learning solution to feature extraction capable of providing hierarchical context information.
This is achieved by employing sparse relative labels indicating relationships of similarity/dissimilarity between image locations.
The nature of these labels results in an almost infinite set of dissimilar examples to choose from. 
We demonstrate how the selection of negative examples during training can be used to modify the feature space and vary it's properties.

To demonstrate the generality of this approach, we apply the proposed features to a multitude of tasks, each requiring different properties. 
This includes disparity estimation, semantic segmentation, self-localisation and SLAM.
In all cases, we show how incorporating SAND features results in better or comparable results to the baseline, whilst requiring little to no additional training.
Code can be found at: {\small \url{https://github.com/jspenmar/SAND_features}}
\end{abstract}

\vspace{-0.3cm}
\section{Introduction}
\vspace{-0.2cm}

\begin{figure}[t]
\begin{center}
\subfloat[Source]{\includegraphics[width=0.5\linewidth]{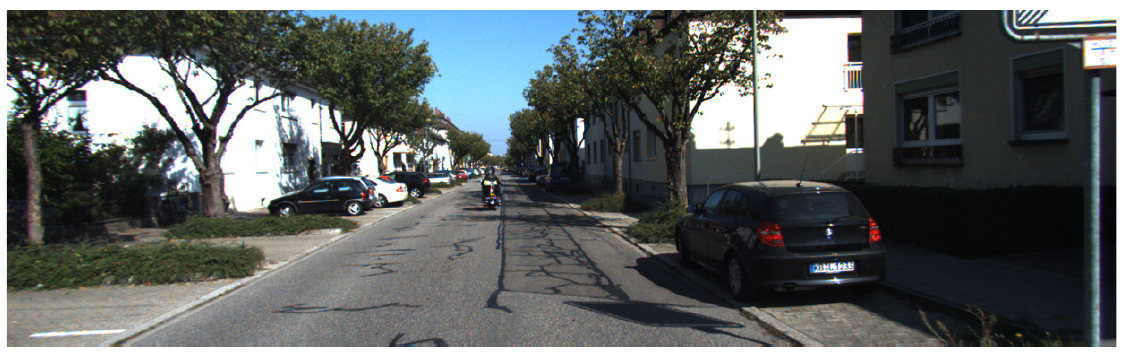}} 
\subfloat[Global]{\includegraphics[width=0.5\linewidth]{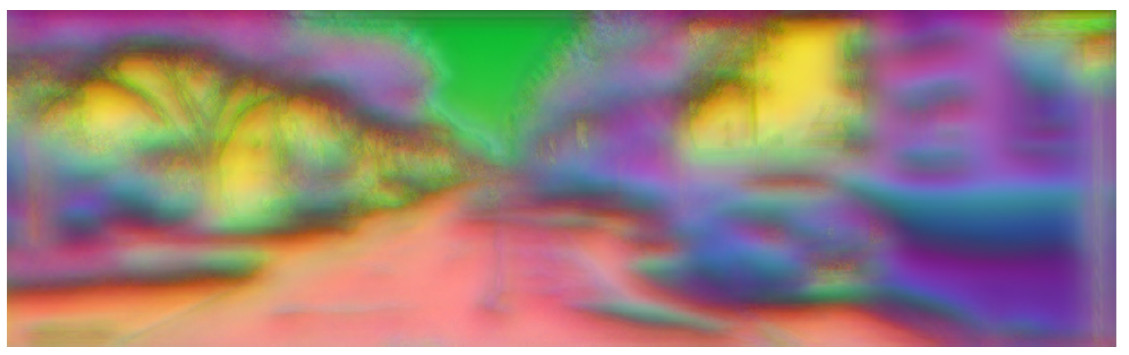}}
\\ \vspace{-0.4cm}
\subfloat[Local]{\includegraphics[width=0.5\linewidth]{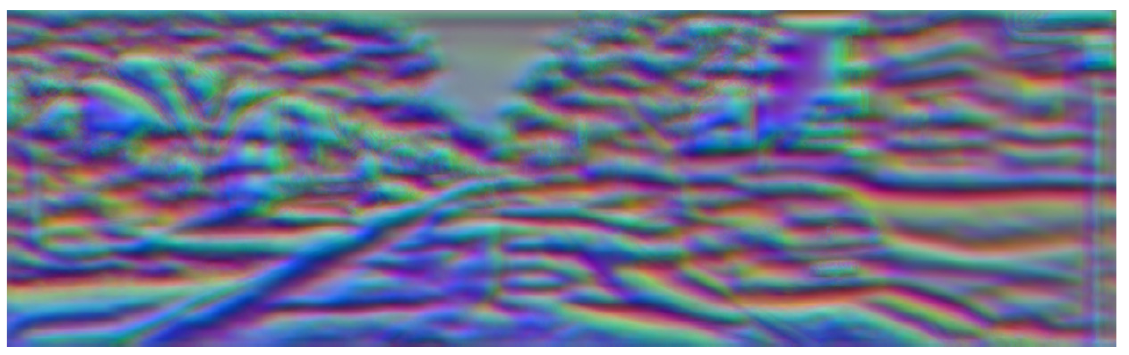}}
\subfloat[Hierarchical]{\includegraphics[width=0.5\linewidth]{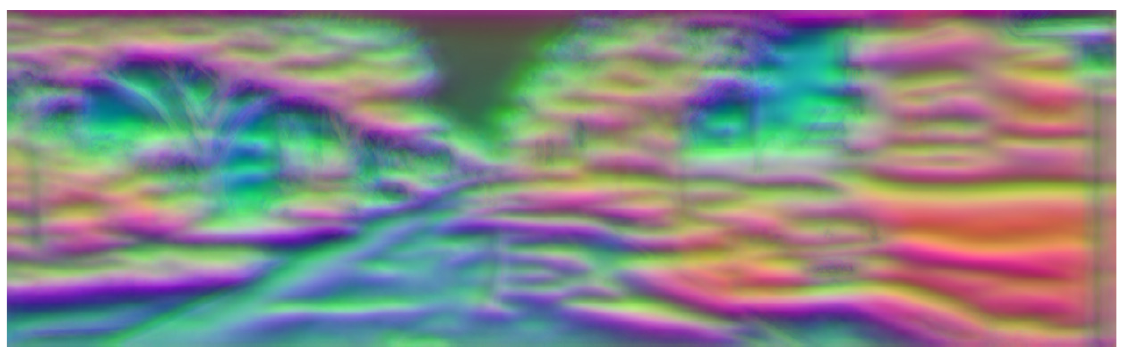}}
\end{center}
\vspace{-0.5cm}
\caption{Visualization of SAND features trained using varying context hierarchies to target specific properties.}
\vspace{-0.6cm}
\label{fig: intro}
\end{figure}

Feature extraction and representation is a fundamental component of most computer vision research.
We propose to learn a feature representation capable of supporting a wide range of computer vision tasks.
Designing such a system proves challenging, as it requires these features to be both unique and capable of generalizing over radical changes in appearances at the pixel-level.
Areas such as \ac{SLAM} or \ac{VO} tend to use feature extraction in an explicit manner \cite{Badino2013, Johnson2008, Kitt2010, Zhang2015}, where hand-crafted sparse features are extracted from pairs of images and matched against each other. 
This requires globally consistent and unique features that are recognisable from wide baselines. 

On the other hand, methods for optical flow \cite{Patel2013} or object tracking \cite{Balasundaram2017} might instead favour locally unique or smooth feature spaces since they tend to require iterative processes over narrow baselines.  
Finally, approaches typically associated with deep learning assume feature extraction to be implicitly included within the learning pipeline. 
End-to-end methods for semantic segmentation \cite{Chen}, disparity estimation \cite{Yang2017} or camera pose regression \cite{Naseer} focus on the learning of implicit ``features'' specific to each task.

Contrary to these approaches, we treat feature extraction as it's own separate deep learning problem. 
By employing sparsely labelled correspondences between pairs of images, we explore approaches to automatically learn dense representations which solve the correspondence problem while exhibiting a range of potential properties. 
In order to learn from this training data, we extend the concept of contrastive loss \cite{Hadsell2006} to pixel-wise non-aligned data.
This results in a fixed set of positive matches from the ground truth correspondences between the images, but leaves an almost infinite range of potential negative samples. 
We show how by carefully targeting specific negatives, the properties of the learned feature representations can be modified to adapt to multiple domains, as shown in Figure \ref{fig: intro}. 
Furthermore, these features can be used in combination with each other to cover a wider range of scenarios.
We refer to this framework as \ac{SAND} features.

Throughout the remainder of this paper we demonstrate the generality of the learned features across several types of computer vision tasks, including stereo disparity estimation, semantic segmentation, self-localisation and \ac{SLAM}.
Disparity estimation and semantic segmentation first combine stereo feature representations to create a 4D cost volume covering all possible disparity levels.
The resulting cost volume is processed in a 3D stacked hourglass network \cite{Chang}, using intermediate supervision and a final upsampling and regression stage.
Self-localisation uses the popular PoseNet \cite{Kendall2015}, replacing the raw input images with our dense 3D feature representation.
Finally, the features are used in a sparse feature matching scenario by replacing ORB/BRIEF features in \ac{SLAM} \cite{Pire2017}.

Our contributions can be summarized as follows:
\begin{enumerate}
	\item We present a methodology for generic feature learning from sparse image correspondences.
	\item Building on ``pixel-wise'' contrastive losses, we demonstrate how targeted negative mining can be used to alter the properties of the learned descriptors and combined into a context hierarchy. 
	\item We explore the uses for the proposed framework in several applications, namely stereo disparity, semantic segmentation, self-localisation and \ac{SLAM}. This leads to better or comparable results in the corresponding baseline with reduced training data and little or no feature finetuning.
\end{enumerate}

\vspace{-0.2cm}
\section{Related Work}
\vspace{-0.2cm}

Traditional approaches to matching with hand-engineered features typically rely on sparse keypoint detection and extraction.
SIFT \cite{Lowe2004} and ORB \cite{Rublee2012}, for example, still remain a popular and effective option in many research areas.
Such is the case with ORB-SLAM \cite{Mur-Artal2015a, Mur-Artal2016} and its variants for \ac{VO} \cite{Forster2017, Zhou2017} or visual object tracking methods, including Sakai \etal \cite{Sakai2015}, Danelljan \etal \cite{Martin2016, Danelljan2016} or Wu \etal \cite{Wu2012a}.
In these cases, only globally discriminative features are required, since the keypoint detector can remove local ambiguity.

As an intermediate step to dense feature learning, some approaches aim to learn both keypoint detection and feature representation. 
Most methods employ hand-crafted feature detectors as a baseline from which to collect data, such as \cite{Altwaijry, Lenc2016}. 
Alternative methods include Salti \etal \cite{Salti2015}, who treat keypoint detection as a binary classification task, and Georgakis \etal \cite{Georgakis}, who instead propose a joint end-to-end detection and extraction network. 

On the other hand, most approaches to dedicated feature learning tend to focus on solving dense correspondence estimation rather than using sparse keypoints.
Early work in this area did not perform explicit feature extraction and instead learns a task specific latent space. 
Such is the case with end-to-end \ac{VO} methods \cite{Wang2017, Wang2018}, camera pose regression \cite{Kendall2015, Brahmbhatt2018} or stereo disparity estimation \cite{Zhang2018}.
Meanwhile, semantic and instance segmentation approaches such as those proposed by Long \etal \cite{Long2015}, Noh \etal \cite{Noh2015} or Wang \etal \cite{Wang2018b} produce a dense representation of the image containing each pixel's class. 
These require dense absolute labels describing specific properties of each pixel.
Despite advances in the annotation tools \cite{Dasiopoulou}, manual checking and refinement still constitutes a significant burden.

Relative labels, which describe the relationships of similarity or dissimilarity between pixels, are much easier to obtain and are available in larger quantities.
Chopra \etal \cite{Chopra2005}, Sun \etal \cite{Sun} and Kang \etal \cite{Kang2018} apply these to face re-identification, which requires learning a discriminative feature space that can generalize over a large amount of unseen data.
As such, these approaches make use of relational learning losses such as contrastive \cite{Hadsell2006} or triplet loss \cite{Schroff2015a}. 
Further work by Yu \etal \cite{Yu} and Ge \etal \cite{Ge} discusses the issues caused by triplet selection bias and provides methods to overcome them. 

As originally presented, these losses don't tackle dense image representation and instead compare holistic image descriptors. 
Schmidt \etal \cite{Schmidt2017} propose a ``pixel-wise'' contrastive loss based on correspondences obtained from KinectFusion \cite{Newcombe} and DynamicFusion \cite{Newcombea}. 
Fathy \etal \cite{Fathy2018a} incorporate an additional matching loss for intermediate layer representations. 
More recently, the contextual loss \cite{Mechrez2018a} has been proposed as a similarity measure for non-aligned feature representations.
In this paper we generalise the concept of ``pixel-wise'' contrastive loss to generic correspondence data and demonstrate how the properties of the learned feature space can be manipulated. 

\vspace{-0.3cm}
\section{SAND Feature Extraction}
\vspace{-0.2cm}

The aim of this work is to provide a high-dimensional feature descriptor for every pixel within an image, capable of describing the context at multiple scales.
We achieve this by employing a pixel-wise contrastive loss in a siamese network architecture.

\begin{figure}[tb!]
\vspace{-0.5cm}
\begin{center}
\includegraphics[width=\linewidth]{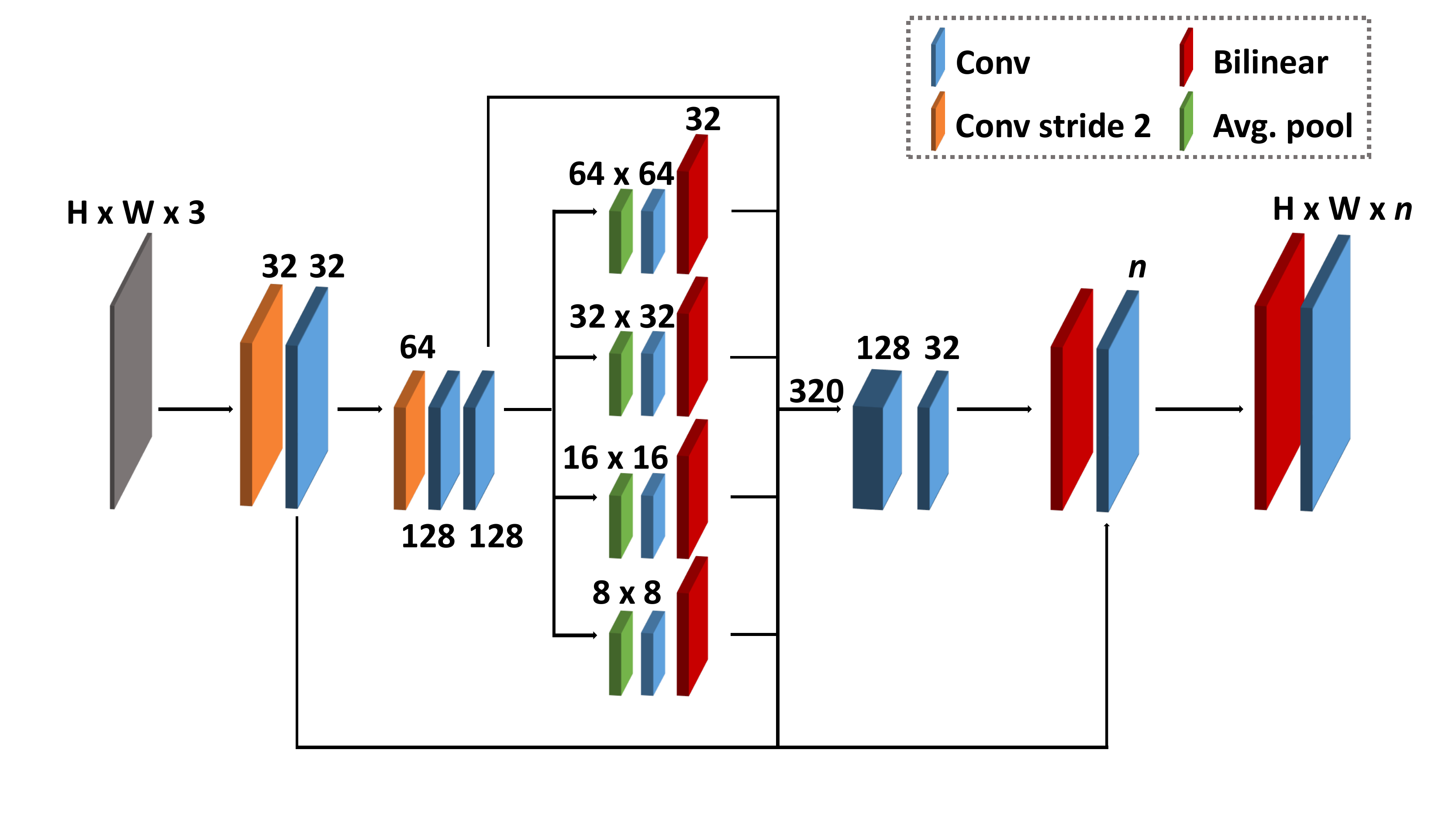}
\end{center}
\vspace{-0.7cm}
\caption{\ac{SAND} architecture trained for dense feature extraction. The initial convolutions are residual blocks, followed by a 4-branch SPP module and multi-stage decoder.}
\vspace{-0.2cm}
\label{fig: net}
\end{figure}

Each branch of the siamese network consists of a series of convolutional residual blocks followed by a \ac{SPP} module, shown in Figure \ref{fig: net}.
The convolution block and base residual blocks serve as the initial feature learning. 
In order to increase the receptive field, the final two residual blocks employ an atrous convolution with dilations of two and four, respectively. 

The \ac{SPP} module is formed by four parallel branches, each with average pooling scales of 8, 16, 32 and 64, respectively. 
Each branch produces a 32D output with a resolution of $(\sfrac{H}{4}, \sfrac{W}{4})$. 
In order to produce the final dense feature map, the resulting block is upsampled in several stages incorporating skip connections and reducing it to the desired number of dimensions, $n$.

Given an input image $I$, it's dense $n$-dimensional feature representation can be obtained by 
\vspace{-0.2cm}
\begin{equation}
	F(\boldsymbol{p}) = \Phi(I(\boldsymbol{p}) | w),
	\vspace{-0.2cm}
\end{equation}
where $\boldsymbol{p}$ represents a 2D point and $\Phi$ represents a \ac{SAND} branch, parametrized by a set of weights $w$. 
$I$ stores RGB colour values, whereas $F$ stores $n$-dimensional feature descriptors, $\Phi: \mathbb{N}^3 \to \mathbb{R}^n$.

\subsection{Pixel-wise Contrastive Loss}
\vspace{-0.2cm}
To train this feature embedding network we build on the ideas presented in \cite{Schmidt2017} and propose a pixel-wise contrastive loss. 
A siamese network with two identical \ac{SAND} branches is trained using this loss to produce dense descriptor maps. 
Given a pair of input points, contrastive loss is defined as
\vspace{-0.2cm}
\begin{equation} \label{eq: contrastive}
	l(y, \p1, \p2) = 
    \begin{cases}
    \frac{1}{2}(d)^2 & \text{if } y = 1 \\
    \frac{1}{2}\{\max(0, m - d)\}^2 & \text{if } y = 0 \\
    0 & otherwise
    \end{cases} \vspace{-0.2cm}
\end{equation}
where $d$ is the euclidean distance of the feature embeddings $||\F1(\p1) - \F2(\p2)||$, $y$ is the label indicating if the pair is a match and $m$ is the margin. 
Intuitively, positive pairs (matching points) should be close in the latent space, while negative pairs (non-matching points) should be separated by at least the margin.

The labels indicating the similarity or dissimilarity can be obtained through a multitude of sources.
In the simplest case, the correspondences are given directly by disparity or optical flow maps. 
If the data is instead given as homogeneous 3D world points $\dot{\boldsymbol{q}}$ in a depth map or pointcloud, these can be projected onto pairs of images.
A set of corresponding pixels can be obtained through
\vspace{-0.3cm}
\begin{equation}
	\boldsymbol{p} = \pi(\dot{\boldsymbol{q}}) = \boldsymbol{K} \boldsymbol{P} \dot{\boldsymbol{q}},
	\vspace{-0.2cm}
\end{equation}
\begin{equation} \label{eq: proj}
	(\C1, \C2) = (\p1, \p2) \ \text{where } \pi_1(\dot{\boldsymbol{q}}) \mapsto \pi_2(\dot{\boldsymbol{q}}),
\end{equation}
where $\pi$ is the projection function parametrized by the corresponding camera's intrinsics $\boldsymbol{K}$ and global pose $\boldsymbol{P}$.

A label mask $\boldsymbol{Y}$ is created indicating if every possible combination of pixels is a positive example, negative example or should be ignored. 
Unlike a traditional siamese network, every input image has many matches, which are not spatially aligned. 
As an extension to (\ref{eq: contrastive}) we obtain
\vspace{-0.2cm}
\begin{equation} \label{eq: full_contrastive}
	L(\boldsymbol{Y}, \F1, \F2) = \sum_{\p1} \sum_{\p2} l(\boldsymbol{Y}(\p1, \p2), \, \p1, \, \p2).
\end{equation}
\vspace{-0.3cm}
%

\subsection{Targeted Negative Mining}
\begin{figure}[t]
\begin{center}
\subfloat[Source]{\includegraphics[width=0.5\linewidth]{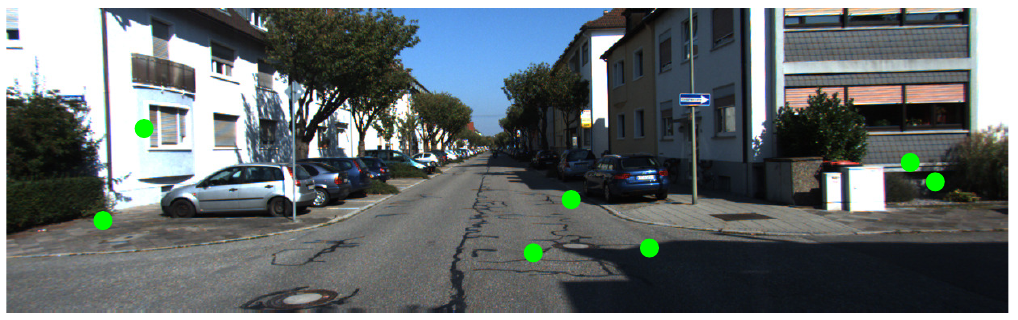}} 
\subfloat[$(0, \infty)$]{\includegraphics[width=0.5\linewidth]{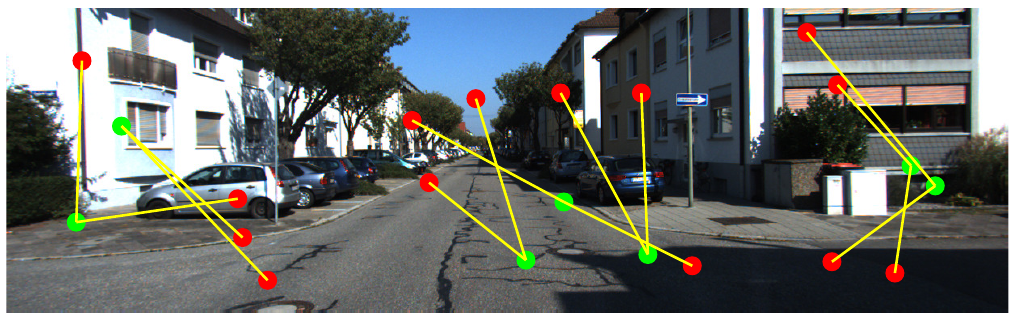}}
\\ \vspace{-0.4cm}
\subfloat[$(0, 25)$]{\includegraphics[width=0.5\linewidth]{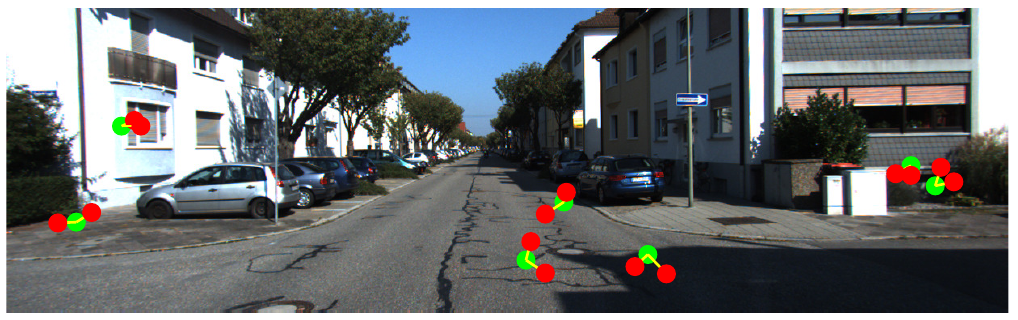}}
\subfloat[$(0, \infty)$ - $(0, 25)$]{\includegraphics[width=0.5\linewidth]{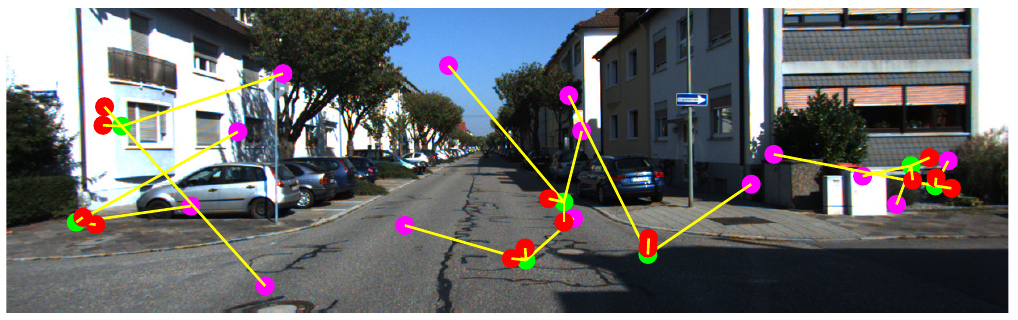}}
\end{center}
\vspace{-0.6cm}
\caption{Effect of $(\alpha, \beta)$ thresholds on the scale information observed by each individual pixel. Large values of $\alpha$ and $\beta$ favour global features, while low $\beta$ values increase local discrimination. \vspace{-0.5cm}}
\label{fig: mining}
\end{figure}

\begin{figure*}[b]
\vspace{-0.2cm}
\begin{center}
\includegraphics[width=1\linewidth]{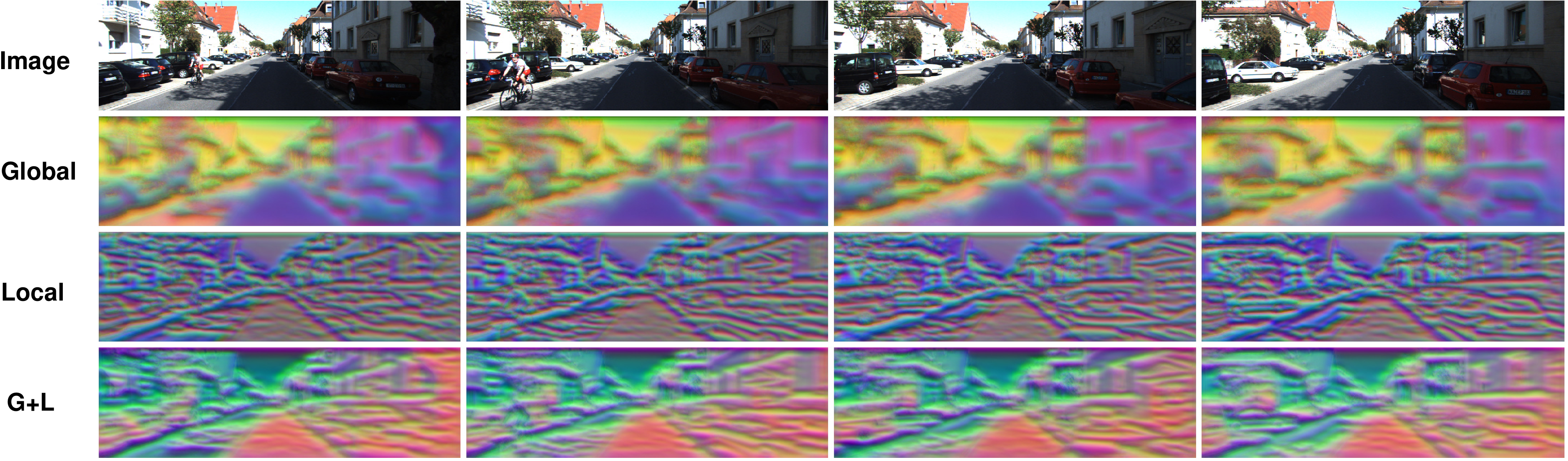}
\end{center}
\vspace{-0.6cm}
\caption{Learned descriptor visualizations for 3D. From top to bottom: source image, \textbf{G}lobal mining, 25 pixel \textbf{L}ocal mining and hierarchical approach. \textbf{L} descriptors show more defined edges and local changes, whereas \textbf{GL} provides a combination of both.\vspace{-0.5cm}}
\label{fig: viz}
\end{figure*}

The label map $\boldsymbol{Y}$ provides the list of similar and dissimilar pairs used during training.
The list of similar pairs is limited by the ground truth correspondences between the input images. 
However, each of these points has $(H \times W) - 1$ potential dissimilar pairs $\hat{\boldsymbol{c}}_2$ to choose from. 
This only increases if we consider all potential dissimilar pairs within a training batch.
For converting 3D ground truth data we can define an equivalent to (\ref{eq: proj}) for negative matches,
\vspace{-0.2cm}
\begin{equation}
\hat{\boldsymbol{c}}_2 \sim \p2 \ \text{where } \pi_1^{-1}(\C1) \nleftrightarrow  \pi_2^{-1}(\p2).
\vspace{-0.2cm}
\end{equation} 
%

It is immediately obvious that it is infeasible to use all available combinations due to computational cost and balancing. 
In the na\"ive case, one can simply select a fixed number of random negative pairs for each point with a ground truth correspondence. 
By selecting a larger number of negative samples, we can better utilise the variability in the available data.
It is also apparent that the resulting highly unbalanced label distributions calls for loss balancing, where the losses attributed to negative samples are inversely weighted according to the total number of pairs selected.  

In practice, uniform random sampling serves to provide globally consistent features.
However, these properties are not ideal for many applications.
By instead intelligently targeting the selection of negative samples we can control the properties of the learned features.

Typically, negative mining consists of selecting hard examples, \ie examples that produce false positives in the network.
Whilst this concept could still be applied within the proposed method, we instead focus on spatial mining strategies, as demonstrated in Figure \ref{fig: mining}.
The proposed mining strategy can be defined as
\vspace{-0.2cm}
\begin{equation}
	\hat{\boldsymbol{c}'}_2 \sim \hat{\boldsymbol{c}}_2 \ \text{where } \alpha < ||\hat{\boldsymbol{c}}_2 - \C2|| < \beta.
	\vspace{-0.2cm}
\end{equation}
In other words, the negative samples are drawn from a region within a radius with lower and higher bounds of $(\alpha, \beta)$, respectively.
As such, this region represents the area in which the features are required to be unique, \ie the \emph{scale} of the features.

For example, narrow baseline stereo requires locally discriminative features.
It is not important for distant regions to be distinct as long as fine details cause measurable changes in the feature embedding.
To encourage this, only samples within a designated radius, \ie a small $\beta$ threshold, should be used as negative pairs. 
On the other hand, global descriptors can be obtained by ignoring nearby samples and selecting negatives exclusively from distant image regions, \ie large $\alpha$ and $\beta=\infty$.

\subsection{Hierarchical Context Aggregation}

It is also possible to benefit from the properties of multiple negative mining strategies simultaneously by ``splitting'' the output feature map and providing each section with different negative sampling strategies. 
For $N_S$ number of mining strategies, $N_C$ represents the number of channels per strategy, $\lfloor \sfrac{n}{N_S} \rfloor$.
As a modification to (\ref{eq: contrastive}), we define the final pixel-level loss as 
\vspace{-0.2cm}
\begin{equation}
\begin{small} \!\!\!\! l(y, \p1, \p2^1 ... \p2^{N_S}\!) \!\! = \!\!
\begin{cases}
\!\! \frac{1}{2} \!\! \sum\limits^{N_S}_{i=1} d^2(i) & \!\!\!\! \text{if } y = 1 \\[10pt]
\!\! \frac{1}{2} \!\! \sum\limits^{N_S}_{i=1} \! \{\max(0, m_i \text{-} d(i))\}^2 & \!\!\!\! \text{if } y = 0 \\[10pt]
0 &  \!\!\!\! otherwise
\end{cases}
\end{small}
\vspace{-0.2cm}
\end{equation}
where $\p2^{i}$ represents a negative sample from strategy $i$ and
\vspace{-0.2cm}
\begin{equation}
d^2(i) =  \sum_{z=iN_C}^{(i+1)N_C} \left(\F1(\p1, z) - \F2(\p2^i, z) \right)^2.
\vspace{-0.2cm}
\end{equation}

This represents a powerful and generic tool that allows us to further adapt to many tasks.
Depending on the problem at hand, we can choose corresponding features scales that best suit the property requirements. 
Furthermore, more complicated tasks or those requiring multiple types of feature can benefit from the appropriate scale hierarchy.
For the purpose of this paper, we will evaluate three main categories: global features, local features and the hierarchical combination of both.


\subsection{Feature Training \& Evaluation}
\textbf{Training.}
In order to obtain the pair correspondences required to train the proposed \ac{SAND} features, we make use of the popular Kitti dataset \cite{Geiger2012}.
Despite evaluating on three of the available Kitti challenges (Odometry, Semantics and Stereo) and the Cambridge Landmarks Dataset, the feature network $\Phi$ is pretrained exclusively on a relatively modest subsection of 700 pairs from the odometry sequence 00. 

Each of these pairs has 10-15 thousand positive correspondences obtained by projecting 3D data onto the images, with 10 negative samples each, generated using the presented mining approaches.
This includes thresholds of $(0, \infty)$ for \textbf{G}lobal descriptors, $(0, 25)$ for \textbf{L}ocal descriptors and the hierarchical combination of both (\textbf{GL}).
Each method is trained for 3, 10 and 32 dimensional feature space variants with a target margin of 0.5.

\textbf{Visualization.}
To begin, a qualitative evaluation of the learned features can be found in Figure \ref{fig: viz}.
This visualisation makes use of the 3D descriptors, as their values can simply be projected onto the RGB color cube. 
The exception to this is \textbf{GL}, which makes use of 6D descriptors reduced to 3D through PCA.
It is immediately apparent how the selected mining process affects the learned feature space. 

When considering small image patches, \textbf{G} descriptors are found to be smooth and consistent, while they are discriminative regarding distant features.
Contrary to this, \textbf{L} shows repeated features across the whole image, but sharp contrasts and edges in their local neighbourhood.
This aligns with the expected response from each mining method.
Finally, \textbf{GL} shows a combination of properties from both previous methods.

\textbf{Distance distributions.}
A series of objective measures is provided through the distribution of positive and negative distances in Table \ref{table: distribs}. 
This includes a similarity measure for positive examples $\mu_{+}$ (lower is better) and a dissimilarity measure for negative examples $\mu_{-}$ (higher is better).
Additionally, the \ac{AUC} measure represents the probability that a randomly chosen negative sample will have a greater distance than the corresponding positive ground truth match.
These studies were carried out for both local (25 pixels radius) and global negative selection strategies.
Additionally, the 32D features were tested with an intermediate (75 pixel radius) and fully combined \textbf{GIL }approach.

From these results, it can be seen that the global approach \textbf{G} performs best in terms of positive correspondence representation, since it minimizes $\mu_{+}$ and maximizes the global \ac{AUC} across all descriptor sizes.
On the other hand, \textbf{L} descriptors provide the best matching performance within the local neighbourhood, but the lowest in the global context. 
Meanwhile, \textbf{I} descriptors provide a compromise between \textbf{G} and \textbf{L}.
Similarly, the combined approach provide an intermediate ground where the distance between all negative samples is maximised and the matching performance at all scales is balanced.
All proposed variants significantly outperform the shown ORB feature baseline.
Finally, it is interesting to note that these properties are preserved across the varying number of dimensions of the learnt feature space, revealing the consistency of the proposed mining strategies.

\begin{table}
\vspace{-0.5cm}
\begin{center}
\begin{tabular}{|c|c||c|c|c|c|c|}
\hline
\multirow{2}{*}{D} & \multirow{2}{*}{Mining} & \multirow{2}{*}{$\mu_{+}$} & \multicolumn{2}{|c|}{Global perf.} &  \multicolumn{2}{|c|}{Local perf.}\\ 
\cline{4-7}
& & & AUC & $\mu_{-}$ & AUC & $\mu_{-}$  \\
\hline\hline
32 & \textbf{ORB} & NA & 85.83 & NA & 84.06 & NA \\
\hline
\multirow{3}{*}{3} & \textbf{G} &  \textbf{0.095} & \textbf{98.62} & 0.951 & 84.70 & 0.300 \\
 & \textbf{L} &  0.147 & 96.05 & 0.628 & \textbf{91.92} & 0.564 \\
 & \textbf{GL} (6D) &  0.181 & 97.86 & \textbf{1.161} & 90.67 & \textbf{0.709} \\
\hline
\multirow{3}{*}{10} & \textbf{G} & \textbf{0.095} & \textbf{99.43} & 0.730 & 86.99 & 0.286 \\
 & \textbf{L} & 0.157 & 98.04 & 0.579 & \textbf{93.57} & 0.510 \\
 & \textbf{GL} & 0.187 & 98.60 & \textbf{1.062} & 91.87 & \textbf{0.678} \\
 \hline
\multirow{5}{*}{32} & \textbf{G} & \textbf{0.093} & \textbf{99.73} & 0.746 & 87.06 & 0.266 \\
 & \textbf{I} & 0.120 & 99.61 & 0.675 & 91.94 & 0.406 \\
 & \textbf{L} & 0.156 & 98.88 & 0.592 & \textbf{94.34} & 0.505 \\
 & \textbf{GL} & 0.183 & 99.28 & 0.996 & 93.34 & 0.642 \\
 & \textbf{GIL} & 0.214 & 98.88 & \textbf{1.217} & 91.97 & \textbf{0.784} \\
\hline
\end{tabular}
\end{center}
\vspace{-0.5cm}
\caption{Feature metrics for varying dimensionality and mining method vs. ORB baseline. \textbf{G}lobal and \textbf{L}ocal provide the best descriptors in their respective areas, while \textbf{GL} and \textbf{GIL} maximise the  negative distance and provides a balanced matching performance.\vspace{-0.7cm}}
\label{table: distribs}
\end{table}

\vspace{-0.5cm}
\section{Feature Matching Cost Volumes} \label{sec: cost}
\vspace{-0.2cm}

Inspired by \cite{Chang}, after performing the initial feature extraction on the stereo images, these are combined in a cost volume $\rho$ by concatenating the left and right features across all possible disparity levels, as defined by
\vspace{-0.2cm}
\begin{equation}
\rho(x, y, \delta, z) = 
\begin{cases}
\boldsymbol{F_1}(x, y, z) & \text{if } z \leq n \\
\boldsymbol{F_2}(x + \delta, y, z) & otherwise
\end{cases},
\vspace{-0.2cm}
\end{equation}
where $n$ corresponds to the dimensionality of the feature maps.  
This results in $\rho(H \times W \times D \times 2n)$, with $D$ representing the levels of disparity.
As such, the cost volume provides a mapping from a 4-dimensional index to a single value, $\rho: \mathbb{N}^4 \to \mathbb{R}$.
 
It is worth noting that this disparity replicated cost volume represents an application agnostic extension of traditional dense feature matching cost volumes \cite{Flynn2016}. 
The following layers are able to produce traditional pixel-wise feature distance maps, but can also perform multi-scale information aggregation and deal with viewpoint variance.
The resulting cost volume is fed to a 3D stacked hourglass network composed of three modules. 
In order to reuse the information learned by previous hourglasses, skip connections are incorporated between corresponding sized layers. 
As a final modification, additional skip connections from the early feature extraction layers are incorporated before the final upsampling and regression stages. 

To illustrate the generality of this system, we exploit the same cost volume and network in two very different tasks.
Stereo disparity estimation represents a traditional application for this kind of approach.
Meanwhile, semantic segmentation has traditionally made use of a single input image.
In order to adapt the network for this purpose, only the final layer is modified to produce an output with the desired number of segmentation classes. 

\vspace{-0.3cm}
\section{Results}
\vspace{-0.2cm}
In addition to the previously mentioned disparity and semantic segmentation, we demonstrate applicability of the proposed \ac{SAND} features in two more areas: self-localisation and \ac{SLAM}.
Each of these areas represents a different computer vision problem with a different set of desired properties.

For instance, stereo disparity represents a narrow baseline matching task and as such may favour local descriptors in order to produce sharp response boundaries.
Meanwhile, semantic segmentation makes use of implicit feature extraction in end-to-end learned representations.
Due to the nature of the problem, feature aggregation and multiple scales should improve performance. 

On the other side of the spectrum, self-localisation emphasizes wide baselines and revisitation, where the global appearance of the scene helps determine the likely location.
In this case, it is crucial to have globally robust features that are invariant to changes in viewpoint and appearance.
Furthermore, the specific method chosen makes use of holistic image representations.

Finally, \ac{SLAM} has similar requirements to self-localisation, where global consistency and viewpoint invariance is crucial to loop closure and drift minimization. 
However, it represents a completely different style of application. 
In this case, revisitation is detected through sparse direct matching rather than an end-to-end learning approach.
Furthermore, the task is particularly demanding of it's features, requiring both wide baseline invariance (mapping) and narrow baseline (\ac{VO}).
As such, it is an ideal use case for the combined feature descriptors.

\vspace{-0.1cm}
\subsection{Disparity Estimation}
\vspace{-0.2cm}

\begin{table}[t!]
\centering
\begin{tabular}{|c|c|c|}
\hline
 Method & Train (\%) & Eval (\%) \\
\hline\hline
Baseline \cite{Chang} & 1.49 & 2.87 \\
\hline
\textbf{10D-G}  & 1.19 & 3.00 \\
\textbf{10D-L} & 1.34 & 2.82\\
\textbf{10D-GL} & 1.16 & 2.91\\
\hline
\textbf{32D-G} & \textbf{1.05} & \textbf{2.65} \\
\textbf{32D-L} & 1.09 & 2.85\\
\textbf{32D-GL} & 1.06 & 2.79\\
\hline
\end{tabular}
\caption{Disparity error on Kitti Stereo train/eval split. With less training, the proposed methods achieves comparable or better performance than the baseline.}
\vspace{-0.4cm}
\label{table: disp}
\end{table}

\begin{figure}[b]
\subfloat[Ground Truth]{\includegraphics[width=1\linewidth]{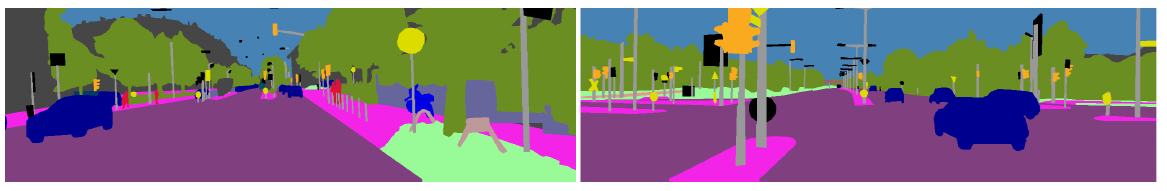}} \\ \vspace{-0.2cm}
\subfloat[Baseline]{\includegraphics[width=1\linewidth]{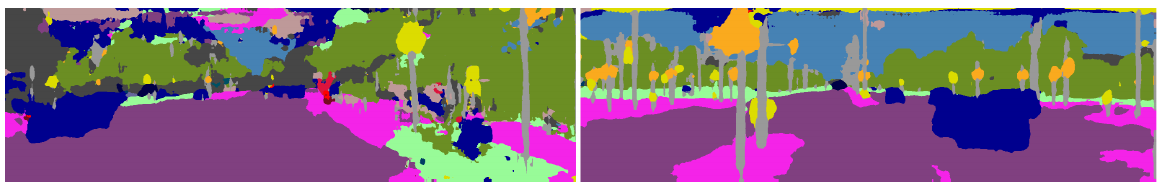}} \\ \vspace{-0.2cm}
\subfloat[\textbf{32D-G-FT}]{\includegraphics[width=1\linewidth]{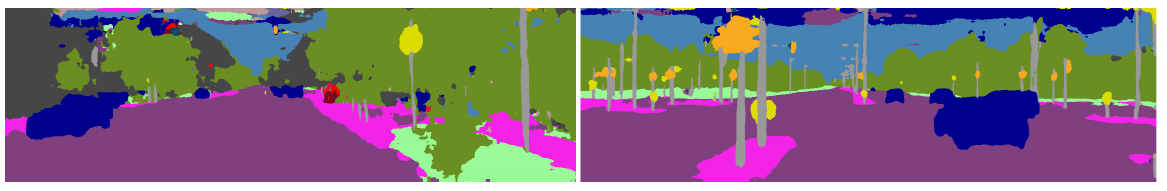}} \\ \vspace{-0.2cm}
\subfloat[\textbf{32D-GL-FT}]{\includegraphics[width=1\linewidth]{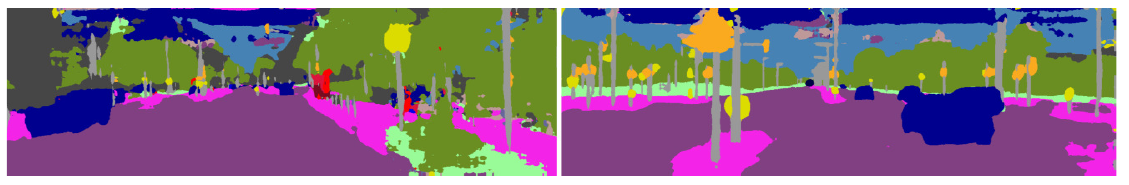}} \\ \vspace{-0.2cm}
\caption{Semantic segmentation visualization for validation set images. The incorporation of SAND features  improves the overall level of detail and consistency of the segmented regions. \vspace{-0.5cm}}
\label{fig: semseg}
\end{figure}

\begin{figure}
\vspace{-0.4cm}
\centering
\subfloat[Baseline]{\includegraphics[width=0.33\linewidth]{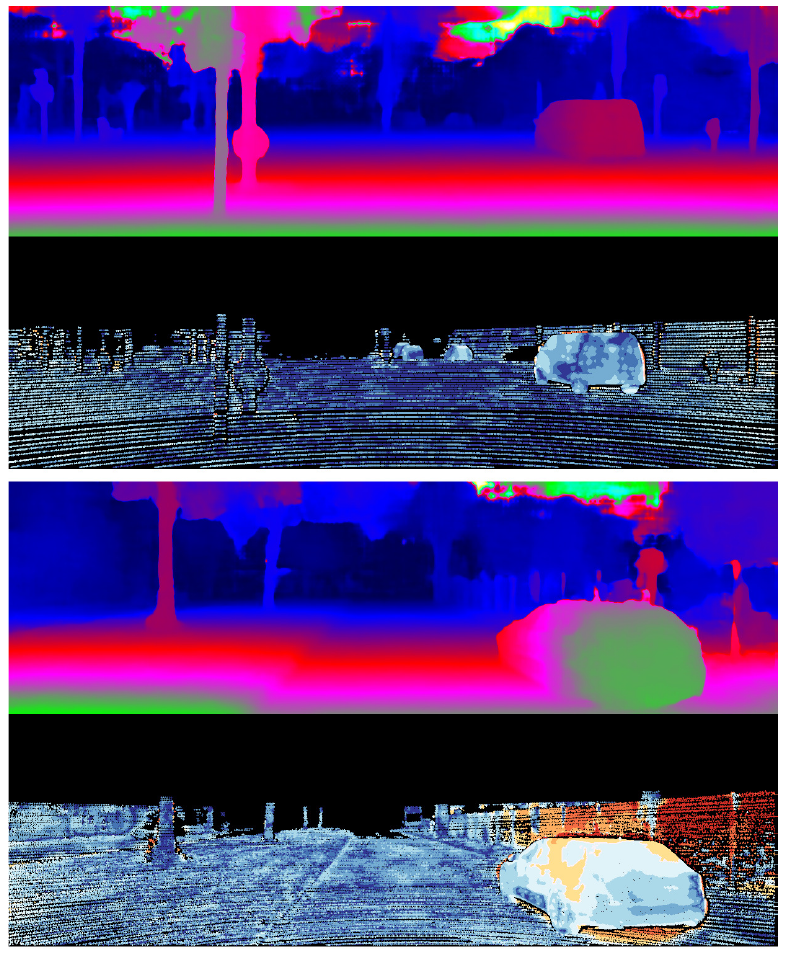}}
\\
\subfloat[\textbf{10D-G}]{\includegraphics[width=0.33\linewidth]{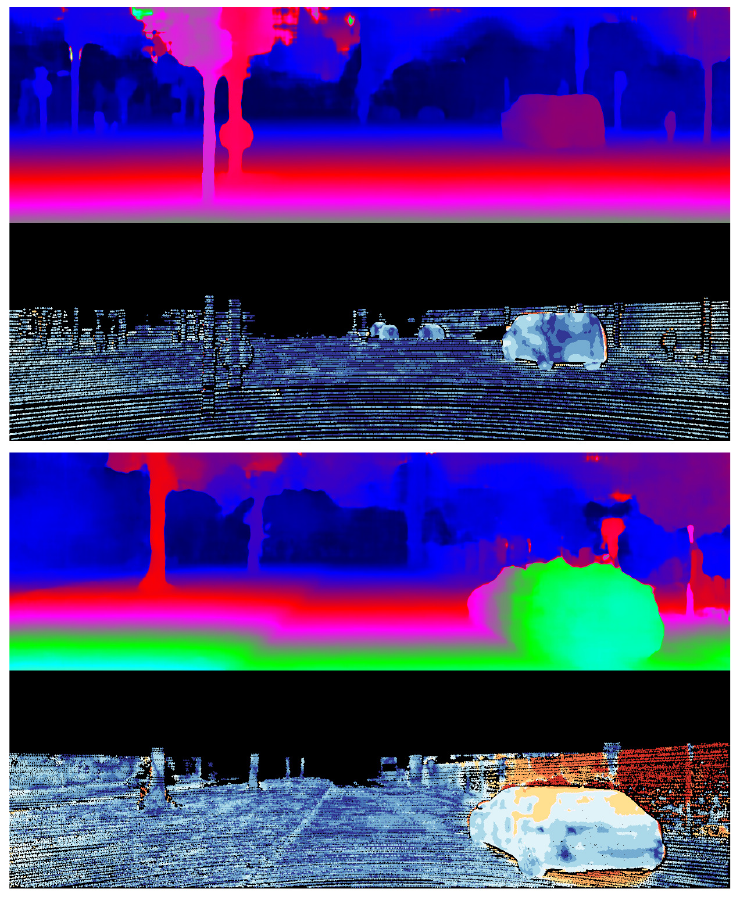}}
\subfloat[\textbf{10D-L}]{\includegraphics[width=0.33\linewidth]{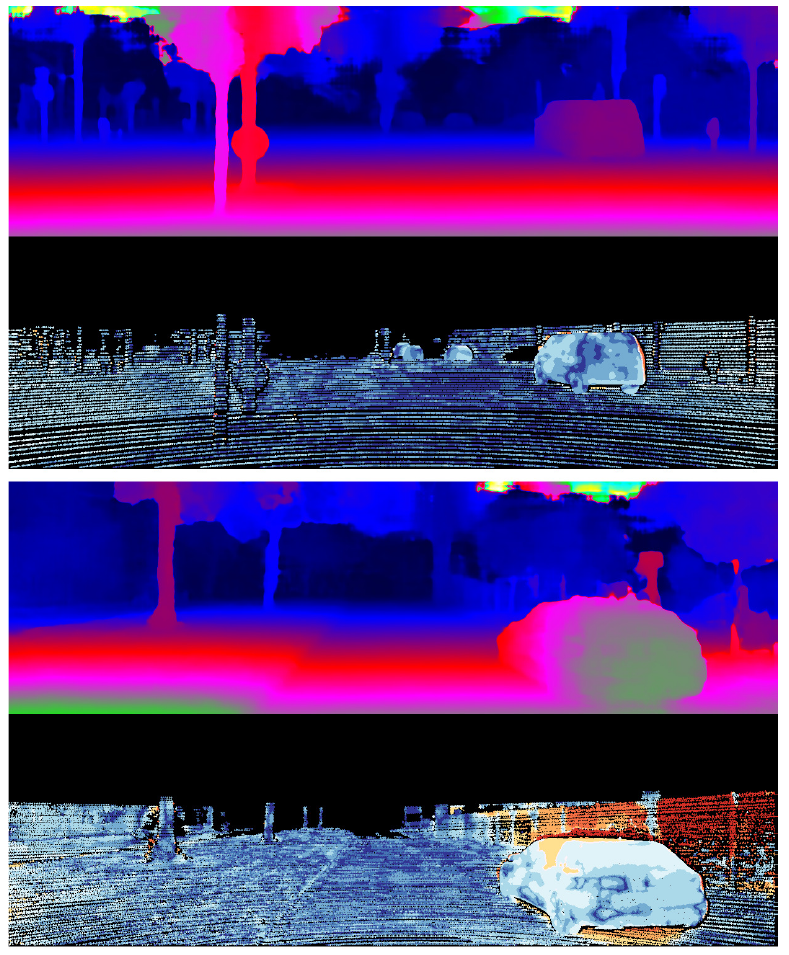}}
\subfloat[\textbf{10D-GL}]{\includegraphics[width=0.33\linewidth]{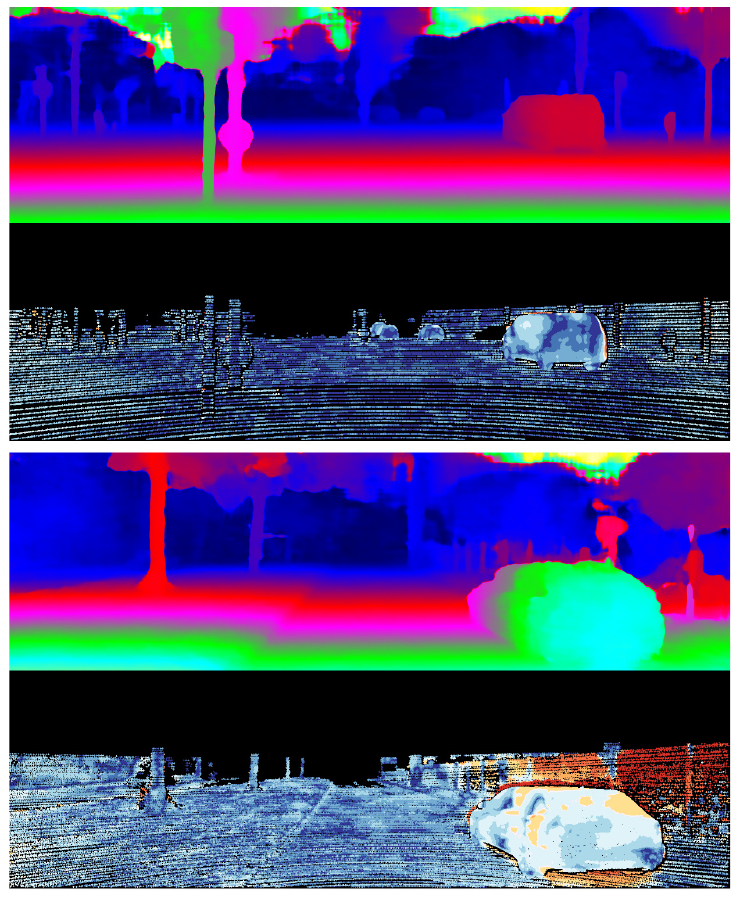}}
\\
\subfloat[\textbf{32D-G}]{\includegraphics[width=0.33\linewidth]{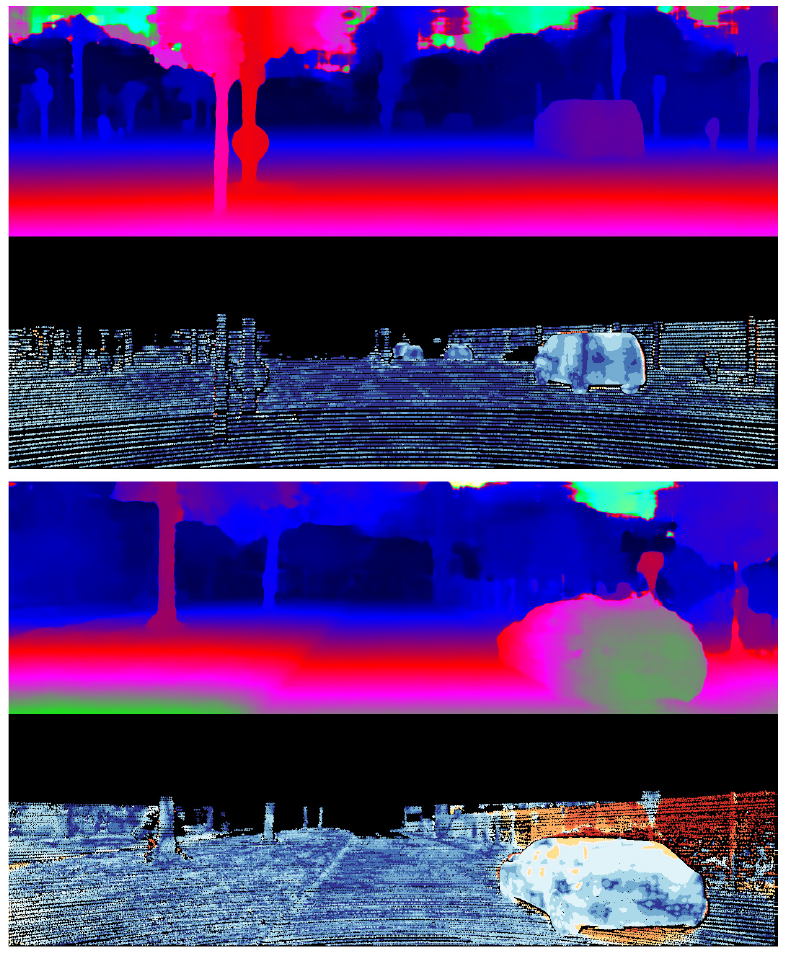}}
\subfloat[\textbf{32D-L}]{\includegraphics[width=0.33\linewidth]{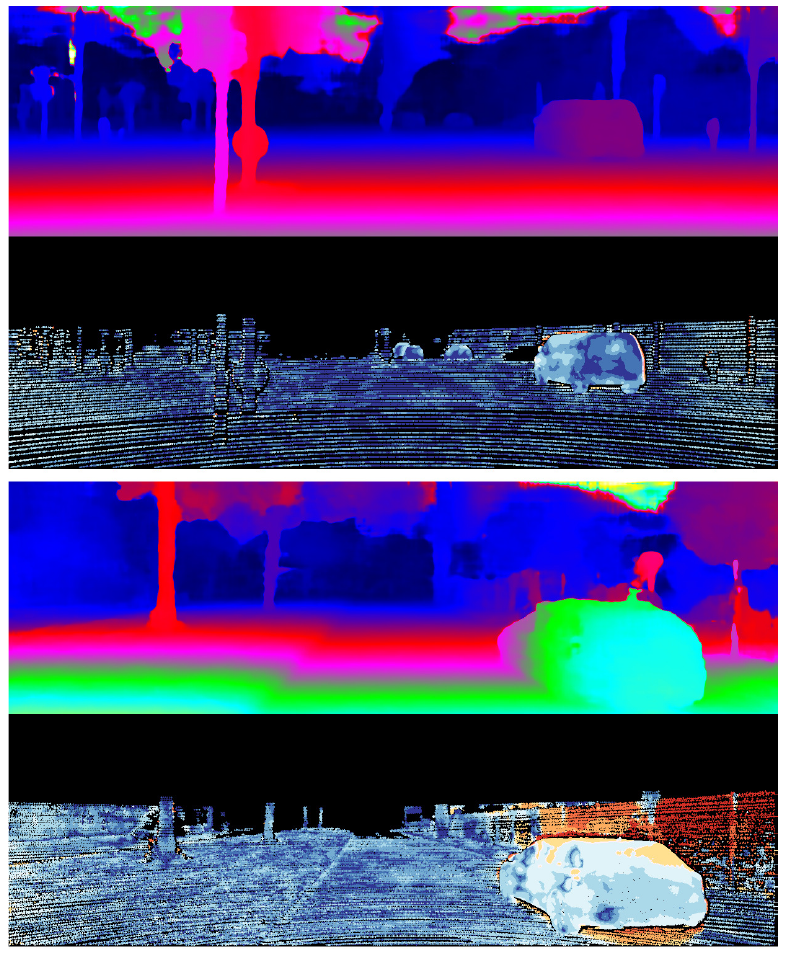}}
\subfloat[\textbf{32D-GL}]{\includegraphics[width=0.33\linewidth]{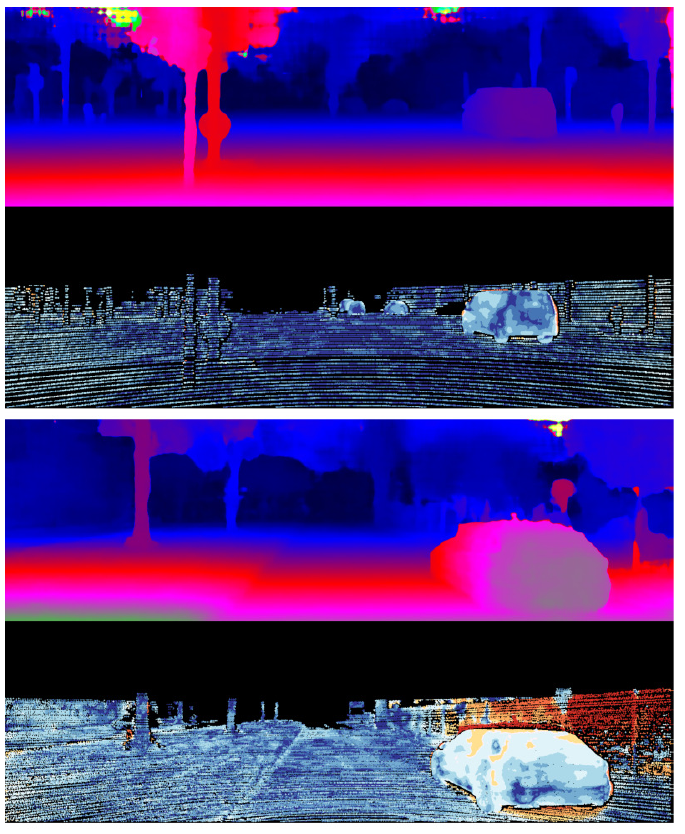}}
\caption{Disparity visualization for two evaluation images (prediction vs. error). The proposed feature representation increases estimation robustness in complicated areas such as the vehicle windows. \vspace{-0.5cm}}
\label{fig: disp}
\end{figure}

Based on the architecture described in Section \ref{sec: cost}, we compare our approach with the implementation in \cite{Chang}.
We compare against the original model trained exclusively on the Kitti Stereo 2015 dataset for 600 epochs. 
Our model fixes the pretrained features for the first 200 epochs and finetunes them at a lower learning rate for 250 epochs. 
The final error metrics on the original train/eval splits from the public Stereo dataset are found in Table \ref{table: disp} (lower is better). 
As seen, with 150 less epochs of training the 10D variant achieves a comparable performance, while the 32D variants provide up to a 30\% reduction in error.
It is interesting to note that \textbf{G} features tend to perform better than local and combined approaches, \textbf{L} and \textbf{GL}.
We theorize that the additional skip connections from the early \ac{SAND} branch make up for any local information required, while the additional global features boost the contextual information. 
Furthermore, a visual comparison for the results in shown in Figure \ref{fig: disp}.
The second and fourth rows provide a visual representation of the error, where red areas indicate larger errors. 
As seen in the bottom row, the proposed method increases the robustness in areas such as the transparent car windows.

%

\vspace{-0.1cm}
\subsection{Semantic Segmentation}
\vspace{-0.2cm}
\begin{table*}
\vspace{-0.5cm}
\begin{center}
\begin{tabular}{|c||c|c|ccccccc|}
\hline
Method & IoU Class & IoU Cat. & Flat & Nature & Object & Sky & Construction & Human & Vehicle \\
\hline\hline
Baseline & 29.3 & 53.8 & 87.1 & 78.1 & 30.1 & \textbf{63.3} & 54.4 & 1.6 & 62.1 \\
\hline
\textbf{32D-G} & 31.1 & 55.8 & 87.3 & 78.5 & 36.0 & 59.8 & 57.5 & \textbf{6.7} & 66.8 \\
\textbf{32D-G-FT} & \textbf{35.4} & \textbf{59.9} & \textbf{88.7} & 83.0 & \textbf{46.7} & 62.7 & \textbf{63.3} & \textbf{6.7} & \textbf{68.1} \\
\hline
\textbf{32D-GL} & 29.4 & 51.7 & 85.1 & 76.6 & 33.8 & 51.8 & 54.4 & 4.3 & 56.3 \\
\textbf{32D-GL-FT} & 33.1 & 56.6 & 87.4 & \textbf{91.5} & 42.6 & 56.7 & 60.4 & 3.9 & 63.7 \\
\hline
\end{tabular}
\end{center}
\vspace{-0.4cm}
\caption{Intersection over Union (\%) measures for class and category average and per-category breakdown. The incorporation of the proposed features results in an increase in accuracy in complicated categories such as Object and Human. \vspace{-0.4cm}} 
\label{table: semseg}
\end{table*}
%

\begin{table*}[b]
\begin{center}
\begin{tabular}{|c||cc|cc|cc|cc|cc|cc|}
\hline
\multirow{2}{*}{Method} & \multicolumn{2}{|c|}{GreatCourt} & \multicolumn{2}{|c|}{KingsCollege} & \multicolumn{2}{|c|}{OldHospital} & \multicolumn{2}{|c|}{ShopFacade} & \multicolumn{2}{|c|}{StMarysChurch} & \multicolumn{2}{|c|}{Street} \\ 
\cline{2-13}

& \textit{P} & \textit{R} & \textit{P} & \textit{R} & \textit{P} & \textit{R} & \textit{P} & \textit{R} & \textit{P} & \textit{R} & \textit{P} & \textit{R} \\
\hline\hline
Baseline & 10.30 & 0.35 & 1.54 & 0.09 & \textbf{3.14} & 0.10 & 2.224 & \textbf{0.19} & \textbf{2.77} & \textbf{0.22} & \textbf{22.60} & 1.01 \\
\hline\hline
\textbf{3D-G} & 12.05 & 0.33 & 2.18 & 0.09 & 4.07 & 0.09 & 2.66 & 0.29 & 4.21 & 0.26 & 36.13 & 1.53\\
\hline
\textbf{32D-G} & 11.46 & 0.30 & 1.62 & 0.09 & 3.30 & 0.11 & 2.20 & 0.25 & 3.67 & 0.23 & 31.92 & 1.24\\
\textbf{32D-G-FT} & \textbf{8.226} & \textbf{0.26} & \textbf{1.52} & \textbf{0.08} & 3.21 & \textbf{0.9} & \textbf{2.01} & 0.22 & 3.16 & \textbf{0.22} & 29.89 & \textbf{0.99}\\
\hline
\end{tabular}
\end{center}
\vspace{-0.4cm}
\caption{Position (m) and Rotation (deg/m) error for baseline PoseNet vs. \ac{SAND} feature variants. \textbf{FT} indicates a variant with finetuned features. The proposed methods outperforms the baseline in half of the sequence in terms of position error and all except one in terms of rotation error. \vspace{-0.4cm}} 
\label{table: posenet}
\end{table*}
Once again, this approach is based on the cost volume presented in Section \ref{sec: cost}, with the final layer producing a 19-class segmentation. 
The presented models are all trained on the Kitti pixel-level semantic segmentation dataset for 600 epochs.
In order to obtain the baseline performance, the stacked hourglass network is trained directly with the input images, whereas the rest use the 32D variants with \textbf{G} and \textbf{LG} learned features.
Unsurprisingly \textbf{L} alone does not contain enough contextual information to converge and therefore is not shown in the following results.

In the case of the proposed methods, two \ac{SAND} variants are trained. The first two fix the features for the first 400 epochs. The remaining two part from these models and finetune the features at a lower learning rate for 200 additional epochs. 

As seen in the results in Table \ref{table: semseg}, the proposed methods significantly outperform the baseline. 
This is especially the case with Human and Object, the more complicated categories where the baseline fails almost completely.
In terms of our features, global features tend to outperform their combined counterpart. 
Again, this shows that this particular task requires more global information in order to determine what objects are present in the scene than exact location information provided by \textbf{L} features.

\vspace{-0.1cm}
\subsection{Self-localisation}
\vspace{-0.2cm}

As previously mentioned, self-localisation is performed using the well known method PoseNet \cite{Kendall2015}. 
While PoseNet has several disadvantages, including additional training for every new scene, it has proven highly successful and serves as an example application requiring holistic image representation. 
The baseline was obtained by training a base ResNet34 architecture as described in \cite{Kendall2015} from scratch with the original dataset images.  
Once again, the proposed method replaces the input images with their respective \ac{SAND} feature representation. 
Both approaches were trained for 100 epochs with a constant learning rate. 
Once again, only versions denoted \textbf{FT} present any additional finetuning to the original pretrained \ac{SAND} features. 

As shown in Table \ref{table: posenet}, the proposed method with 32D finetuned features generally outperforms the baseline. 
This contains errors for the regressed position, measured in meters from the ground truth, and rotation representing the orientation of the camera.
As expected, increasing the dimensionality of the representation (3 vs. 32) increases the final accuracy, as does finetuning the learnt representations. 

Most notably it performs well in sequences like GreatCourt, KingsCollege or ShopFacade.
We theorize that this is due to the distinctive features and shapes of the buildings, which allows for a  more robust representation. 
However, the approach tends to perform worse in sequences containing similar or repeating surroundings, such as the Street sequence. 
This represent a complicated environment for the proposed features in the context of PoseNet, since the global representation can't be reliably correlated with the exact position without additional information.

\vspace{-0.2cm}
\subsection{SLAM}
\vspace{-0.2cm}
\begin{table*}
\vspace{-0.5cm}
\centering
\begin{center}
\begin{tabular}{|c||*{11}{cc|}}
\hline
\multirow{2}{*}{Method} & \multicolumn{2}{|c|}{00} & \multicolumn{2}{|c|}{02} & \multicolumn{2}{|c|}{03} & \multicolumn{2}{|c|}{04} & \multicolumn{2}{|c|}{05} \\
\cline{2-11}
& APE & RPE & APE & RPE & APE & RPE & APE & RPE & APE & RPE \\
\hline\hline
Baseline & 5.63 & 0.21 & \textbf{8.99} & \textbf{0.28} & 6.39 & 0.05 & \textbf{0.69} & \textbf{0.04} & 2.35 & 0.12 \\
\textbf{32D-G} & 13.09 & 0.21 & 41.65 & 0.36 & 6.00 & 0.08 & 6.43 & 0.13 & 6.59 & 0.16 \\
\textbf{32D-L} & 5.99 & 0.21 & 9.83 & 0.29 & 4.40 & \textbf{0.04} & 1.13 & 0.05 & 2.37 & 0.12 \\
\textbf{32D-GL} & \textbf{4.84} & \textbf{0.20} & 9.66 & 0.29 & \textbf{3.69} & \textbf{0.04} & 1.35 & 0.05 & \textbf{1.93} & \textbf{0.11} \\
\hline\hline
Method & \multicolumn{2}{|c|}{06} & \multicolumn{2}{|c|}{07} & \multicolumn{2}{|c|}{08} & \multicolumn{2}{|c|}{09} & \multicolumn{2}{|c|}{10} \\
\hline \hline
Baseline & 3.78 & 0.09 & 1.10 & \textbf{0.19} & \textbf{4.19} & \textbf{0.13} & 5.77 & 0.43 & 2.06 & \textbf{0.28} \\
\textbf{32D-G} & 9.10 & 0.13 & 2.05 & 0.21 & 15.40 & 0.17 & 11.50 & 0.45 & 18.25 & 0.35 \\
\textbf{32D-L} & 2.54 & 0.09 & \textbf{0.88} & \textbf{0.19} & 5.26 & \textbf{0.13} & 6.25 & \textbf{0.42} & 2.03 & 0.30 \\
\textbf{32D-GL} & \textbf{2.00} & \textbf{0.08} & 0.96 & \textbf{0.19} & 6.00 & \textbf{0.13} & \textbf{5.48} & \textbf{0.42} & \textbf{1.36} & 0.29 \\
\hline
\end{tabular}
\end{center}
\vspace{-0.4cm}
\caption{Absolute and relative pose error (lower is better) breakdown for all public Kitti odometry sequences, except 01. APE represents aligned trajectory absolute distance error, while RPE represents motion estimation error. On average, \textbf{32D-GL} provides the best results, with comparable performance from \textbf{32D-L}. \vspace{-0.4cm}}
\label{table: slam}
\end{table*}

All previous areas of work explore the use of our features in a deep learning environment, where the dense feature representations are used.
This set of experiments instead focuses on their use in a sparse matching domain with explicit feature extraction.
The learned features serve as a direct replacement for hand-engineered features.
The baseline \ac{SLAM} system used is an implementation for S-PTAM \cite{Pire2017}. 
This system makes use of ORB descriptors to estimate \ac{VO} and create the environment maps. 

We perform no additional training or adaptation of our features, or any other part of the pipeline for this task. We simply drop our features into the architecture that was built around ORB.
It is worth emphasising that we also do not aggregate our features over a local patch. Instead we rely on the feature extraction network to have already encoded all relevant contextual information in the pixel's descriptor.

\begin{figure}[b!]
\vspace{-0.5cm}
\begin{center}
\subfloat{\includegraphics[width=0.5\linewidth]{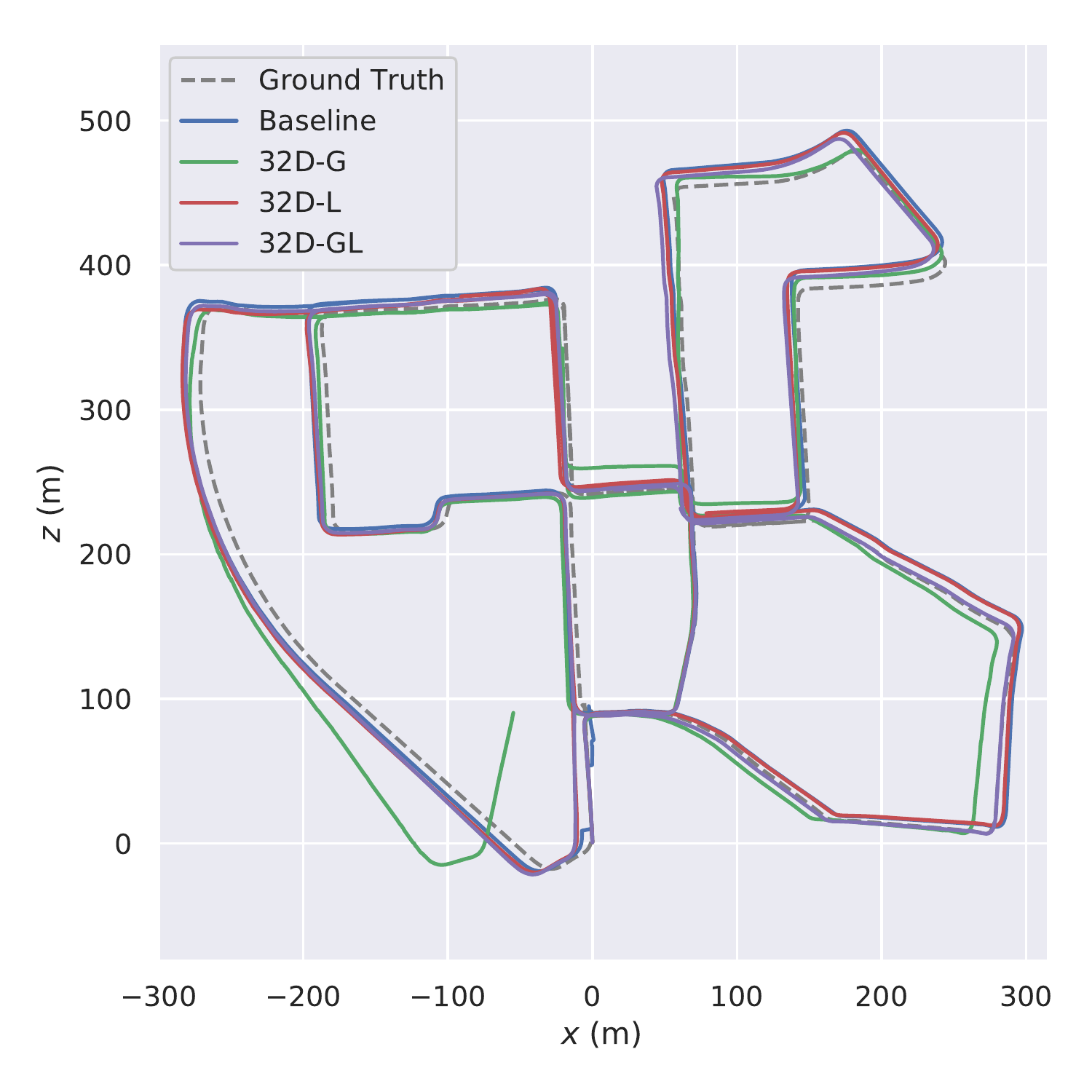}}
\subfloat{\includegraphics[width=0.5\linewidth]{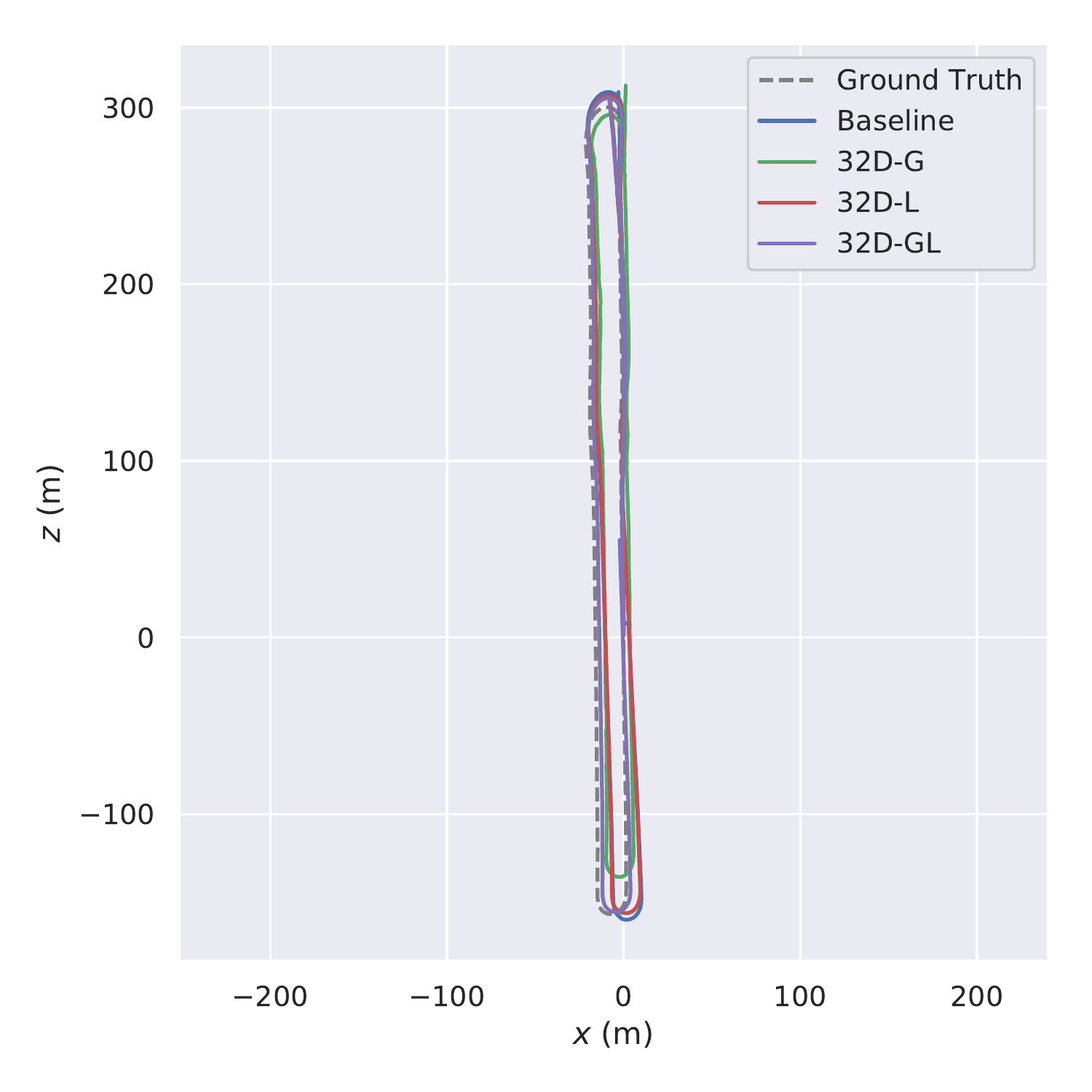}}
\\
\subfloat{\includegraphics[width=0.5\linewidth]{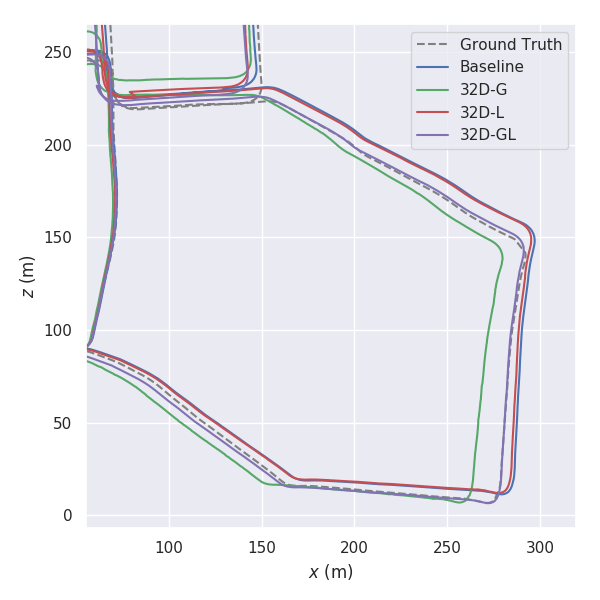}}
\subfloat{\includegraphics[width=0.5\linewidth]{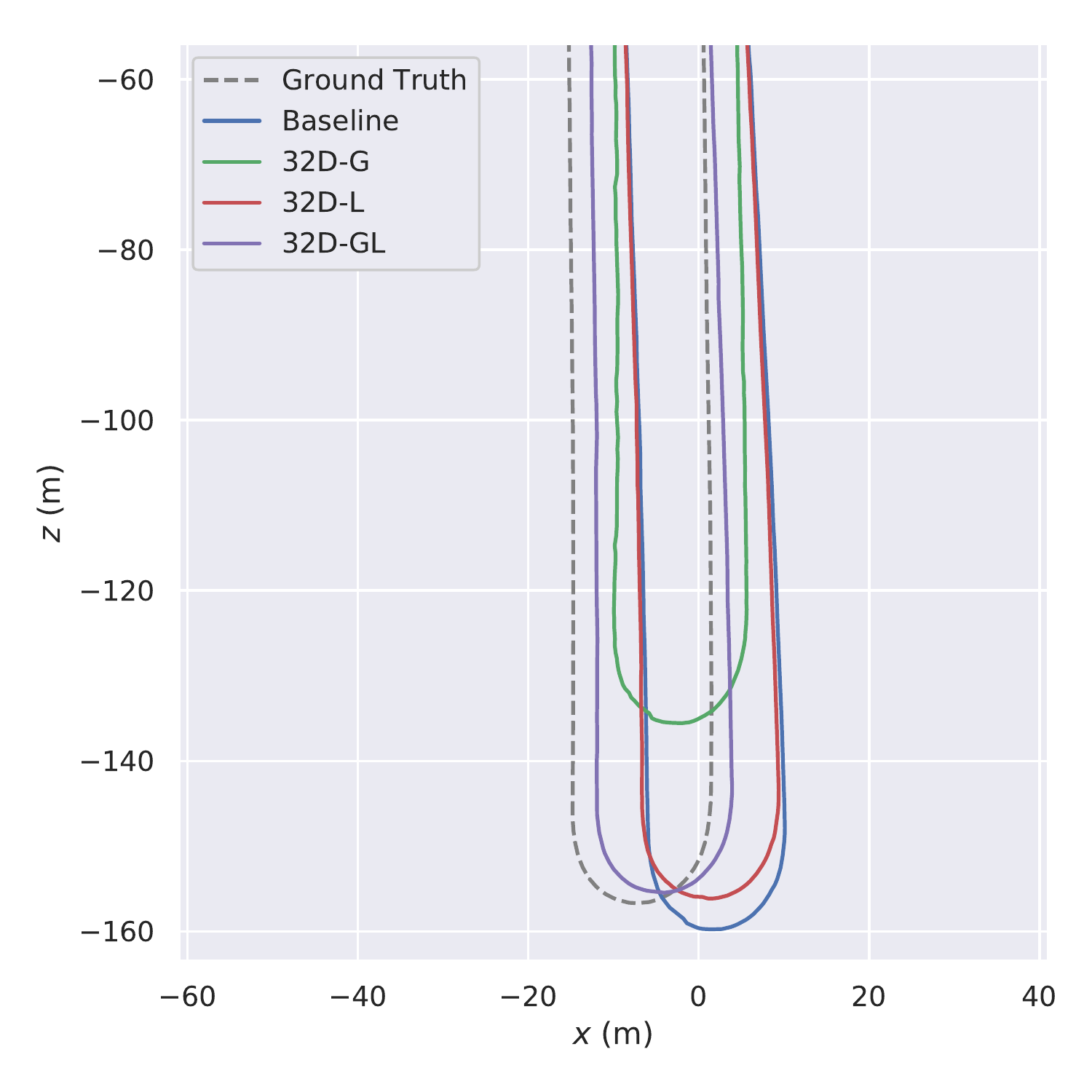}}
\caption{Kitti odometry trajectory predictions for varying \ac{SAND} features vs. baseline. Top row shows two full sequences, with zoomed details in the bottom row. The hierarchical approach \textbf{GL} provides both robust motion and drift correction.\vspace{-0.5cm}}
\label{fig: slam}
\end{center}
\end{figure}

A visual comparison between the predicted trajectories for two Kitti odometry sequences can be found in Figure \ref{fig: slam}.
As seen, the proposed method follows the ground truth more closely and presents less drift. 
In turn, this shows that our features are generally robust to revisitations and are viewpoint invariant.

Additionally, the average absolute and relative pose errors for the available Kitti sequences are shown in Table \ref{table: slam}.
These measures represent the absolute distance between the aligned trajectory poses and the error in the predicted motion, respectively. 
In this application, it can be seen how the system greatly benefits for the hierarchical aggregation learning approach.
This is due to \ac{SLAM} requiring two different sets of features.
In order to estimate the motion of the agent in a narrow baseline, the system requires locally discriminative features.
On the other hand, loop closure detection and map creation requires globally consistent features. 
This is refected in the results, where \textbf{G} consistently drifts more than \textbf{L} (higher RPE) and \textbf{GL} provides better absolute pose (lower APE).

\vspace{-0.2cm}
\section{Conclusions \& Future Work}
\vspace{-0.2cm}
We have presented \ac{SAND}, a novel method for dense feature descriptor learning with a pixel-wise contrastive loss.
By using sparsely labelled data from a fraction of the available training data we demonstrate that it is possible to learn generic feature representations.
While other methods employ hard negative mining as a way to increase robustness, we instead develop a generic contrastive loss framework allowing us to modify and manipulate the learned feature space. 
This results in a hierarchical aggregation of contextual information visible to each pixel throughout training. 

In order to demonstrate the generality and applicability of this approach, we evaluate it on a series of different computer vision applications each requiring different feature properties. 
This ranges from dense and sparse correlation detection to holistic image description and pixel-wise classification.
In all cases \ac{SAND} features are shown to outperform the original baselines.

We hope this is a useful tool for most areas of computer vision research by providing easier to use features requiring less or no training.
Further work in this area could include exploring additional desirable properties for the learnt features spaces and the application of these to novel tasks. 
Additionally, in order to increase the generality of these features they can be trained with much larger datasets containing a larger variety of environments, such as indoor scenes or seasonal changes. 

\subsection*{Acknowledgements}
\vspace{-0.3cm}
This work was funded by the EPSRC under grant agreement (EP/R512217/1). 
We would also like to thank NVIDIA Corporation for their Titan Xp GPU grant.
\vspace{-0.4cm}

{\small
\bibliographystyle{ieee}

}

\end{document}